\documentclass[twoside]{article}

\usepackage{aistats2025}

\usepackage{microtype}
\usepackage{graphicx}
\usepackage{svg}
\usepackage{subcaption}
\usepackage{booktabs} 
\usepackage{hyperref}
\usepackage{amssymb}
\usepackage{amsmath}
\usepackage{commath}
\usepackage{graphicx,bm}
\usepackage{verbatim}
\usepackage{csquotes}
\usepackage{multirow}
\usepackage[round]{natbib}

\begin{document}

%

%

\twocolumn[

\aistatstitle{Multi-level Supervised Contrastive Learning}

\aistatsauthor{Naghmeh Ghanooni$^{1}$ \And  
    Barbod Pajoum$^{1}$\And  
    Harshit Rawal$^{1}$ \AND
    Sophie Fellenz$^{1}$ \And  
    Vo Nguyen Le Duy$^{2}$ \And  
    Marius Kloft$^{1}$
    }
 \vspace{10pt}
\aistatsaddress{$^{1}$RPTU Kaiserslautern-Landau \And $^{2}$RIKEN, Japan}]

\begin{abstract}
Contrastive learning is a well-established paradigm in representation learning. The standard framework of contrastive learning minimizes the distance between \enquote{similar} instances and maximizes the distance between dissimilar ones in the projection space, disregarding the various aspects of similarity that can exist between two samples. Current methods rely on a single projection head, which fails to capture the full complexity of different aspects of a sample, leading to sub-optimal performance, especially in scenarios with limited training data.
In this paper, we present a novel supervised contrastive learning method in a unified framework called multi-level contrastive learning (MLCL), that can be applied to both multi-label and hierarchical classification tasks. The key strength of the proposed method is the ability to capture similarities between samples across different labels and/or hierarchies using multiple projection heads. Extensive experiments on text and image datasets demonstrate that the proposed approach outperforms state-of-the-art contrastive learning methods. 
\end{abstract}

\section{Introduction}
\label{introduction}

Contrastive learning is a key framework in representation learning, with the primary objective of learning adjacent representations for \enquote{similar} samples. One widely adopted approach is supervised contrastive learning \citep{khosla2020supervised}, known as SupCon, where similarity is defined based on groundtruth labels. In SupCon, the model learns representations by bringing samples from the same class closer in the representation space, while distancing those from different classes. However, it relies on a single label per sample to define similarity, posing limitations for multi-label classification and datasets with hierarchical class structures. Existing approaches typically address these challenges separately, using specific loss functions \citep{zhang2022use} tailored for hierarchical structures or learning label-level embeddings \citep{dao2021contrast} for multi-label datasets \citep{zhang2024multi}. In contrast, our method introduces a unified framework called multi-level contrastive learning, which effectively captures both multi-label and hierarchical aspects of data within a single representation space.

\begin{figure*}
\subfloat[\label{fig:text-rep}]{%
  \includegraphics[width=.49\linewidth]{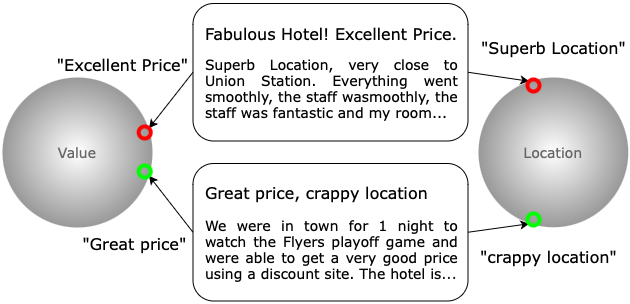}%
}
\hfill
\subfloat[\label{fig:img-rep}]{%
  \includegraphics[width=.49\linewidth]{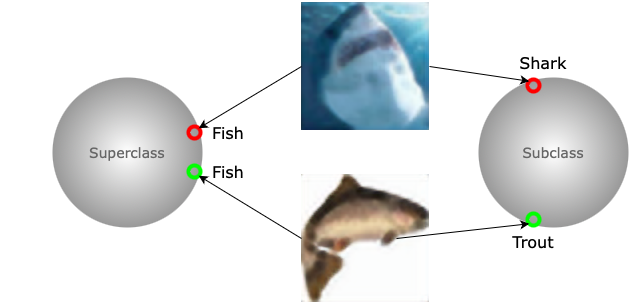}%
}
\caption{An illustration of two projection spaces using different examples: a) TripAdvisor reviews and b) CIFAR-100. Similarity can be defined across various dimensions, and a single projection space is insufficient to capture the full spectrum of feature levels.}
\end{figure*}

In this work, we incorporate multiple projection heads into the standard contrastive learning architecture, each tailored to learn representations from a specific aspect of the instance. These projection heads are connected to the encoder to capture different levels of representation. To enhance the effect of the projection heads, a temperature hyperparameter is used to control class discrimination at each level in the projection space, as the concept of similarity differs across levels. By tuning the temperature, the model focuses on \enquote{hard negatives} at the relevant level while reducing their impact on others. This significantly improves performance, particularly in scenarios with limited training samples. 

The main contributions are summarized as follows:
\begin{itemize}
    \item This paper introduces a novel unified framework that combines hierarchical and multi-label settings into a single multi-level approach.
    \item We propose an end-to-end architecture that generalizes standard contrastive learning through multiple projection heads. 
    \item These projection heads capture label-specific similarities and class hierarchies or serve as a regularizer to prevent overfitting.
    \item We validate the effectiveness of the proposed method on multi-label and hierarchical classification tasks within the domains of natural language processing (NLP) and computer vision.
\end{itemize}

\section{Related Work}
\label{relatedwork}
\paragraph{Contrastive learning}
Learning a high-quality representation of the target data that can be transferred to various downstream tasks is a major challenge in both 
Natural Language Processing(NLP) and computer vision. Contrastive learning is a powerful technique in representation learning. SimCLR~\citep{chen2020simple} introduced a novel framework for acquiring high-quality representations from unlabeled images, achieving state-of-the-art results in the field. Its popularity stems from its simplicity, as it operates without the need for specialized architectures \citep{bachman2019learning, henaff2020data} or memory banks \citep{he2020momentum, wu2018unsupervised}, while demonstrating remarkable performance across diverse downstream tasks. SimCLR discriminates positive and negative pairs in the projection space, and large batch sizes with many negative pairs and various data augmentations are required for optimal performance. 
Several subsequent papers attempted to improve SimCLR’s performance and overcome its limitations. For example, \citet{chuang2020debiased} presents a debiased contrastive loss that corrects the sampling bias of negative examples, while BYOL \citep{grill2020bootstrap} relies only on positive samples for training and is more robust to batch size and data augmentations. Furthermore, \citet{kalantidis2020hard} emphasizes the importance of hard negatives in contrastive learning and proposes a hard negative mixing strategy that improves performance and avoids the need for large batch sizes and memory banks \citep{mocov2}. Recently, \citet{wang2023adaptive} introduced an adaptive multiple projection head mechanism for self-supervised contrastive learning models to address better intra- and inter-sample similarities between different views of the samples. They also incorporate an adaptive temperature mechanism to re-weight each similarity pair. In contrast, our approach employs multiple projection heads to capture different aspects of the samples within a supervised framework, making it suitable for downstream tasks such as multi-label and hierarchical classifications.

\begin{figure*}
\subfloat[Two projection heads $g_1$ and $g_2$ are attached to the encoder's output to capture subclass and superclass similarities respectively.\label{fig:img-arch}]{%
  \includegraphics[height=4cm,width=.49\linewidth]{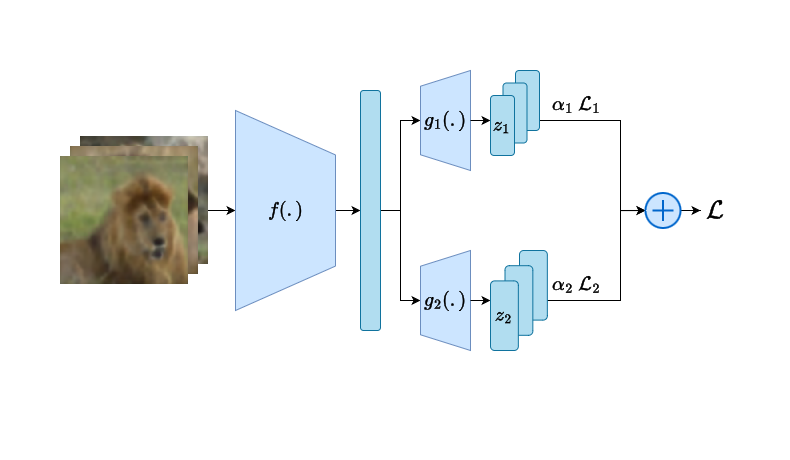}%
}
\hfill
\subfloat[We employ an individual projection head for each label and an additional global head to capture the overall similarities between the two reviews. \label{fig:text-arch}]{%
  \includegraphics[height=4cm,width=.49\linewidth]{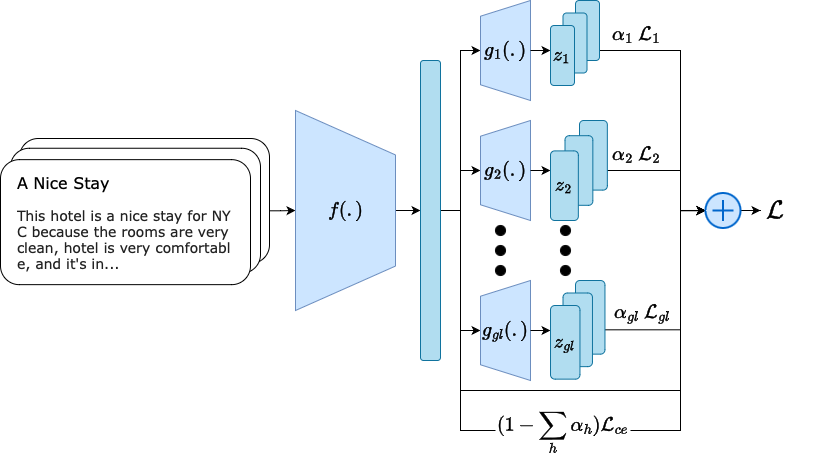}%
}
\caption{
Our proposed architecture with multiple projection heads for a) hierarchical classification and b) multi-label classification. 
The final loss is computed as a weighted sum of the losses from each head, with an additional cross-entropy loss applied in the case of text classification.
}
\end{figure*}

\paragraph{Supervised Contrastive Learning}
The self-supervised contrastive learning approach can be extended to supervised settings with minimal modifications. SupCon~\citep{khosla2020supervised} developed a supervised contrastive learning method that leverages labels by modifying the contrastive loss function. Instead of comparing each sample with all other samples, as in SimCLR, SupCon pulls samples from the same class closer and pushes samples from different classes farther apart. Several studies have aimed to enhance SupCon. For example, \citet{barbano2022unbiased} focused on controlling the minimal distance between positive and negative samples and introduced a new debiasing regularization loss. SupCon suffers from imbalanced data distribution, leading to poor uniformity in the feature distribution on the hypersphere. \citet{li2022targeted} addressed this issue by proposing a targeted supervised contrastive learning loss to maintain a uniform feature distribution even with long-tail distribution data. Additionally, \citet{chen2022perfectly} suggested a weighted class-conditional InfoNCE loss to enhance the robustness of SupCon. Furthermore, certain methods leverage contrastive representation to learn from noisy labels \citep{ciortan2021framework, ghosh2021contrastive, li2021learning, li2020mopro, xue2022investigating, yi2022learning}.  While these approaches indicate the suitability of contrastive learning in the supervised setting, they do not provide the possibility of contrasting multiple aspects.

\paragraph{Multi-label and Hierarchical Contrastive Learning}
Recent research has extended supervised contrastive learning (SCL) to multi-label contrastive learning \citep{dao2021multi, malkinski2022multi, dao2021contrast, zaigrajewcontrastive, zhang2024multi}. These extensions aim to bring samples with shared labels closer together while separating those with different labels. For instance, \citet{zaigrajewcontrastive} utilized the Jaccard similarity measure to assess label overlap between sample pairs and \citet{zhang2024multi} introduced multiple positive sets to address this problem. Meanwhile, \citet{sajedi2023end} introduced a kernel-based approach that transforms embedded features into a mixture of exponential kernels in Gaussian RKHS for multi-label contrastive learning. Additionally, \citet{dao2021contrast} proposed a module that learns multiple label-level representations using multiple attention blocks. However, the method by \citet{dao2021contrast}, which trains label-specific features, involves two-stage training and produces task-specific representations not universally applicable across all downstream tasks.  In contrast, our approach considers comprehensive similarity aspects among samples beyond mere label overlap, ensuring applicability across a wide range of classification tasks. Incorporating the class hierarchies, \citet{landrieu2021leveraging} (
Guided) integrate class distance into a prototypical network to model the hierarchical class structure and \citet{zhang2022use} (HiMulConE) introduced a hierarchical framework for multi-label representation learning with a new loss function penalizing sample distance in label hierarchies. Existing approaches often overlook diverse similarity perspectives, tailoring solutions to specific scenarios rather than broader applications.

\section{Approach}
\label{approach}

\subsection{Problem Formulation}
\label{sec:formulation}
For simplicity, we adopt a unified formulation for both multi-label and hierarchical classification. In each training iteration, a batch of randomly sampled inputs is given to the network, where each input is associated with \( L \) levels: \(\{x_k, y_k^l\}\), for \( k \in \{1, \dots, B\} \) and \( l \in \{1, \dots, L\} \). In multi-label classification, \( y_k^l \) corresponds to the labels assigned to sample \( x_k \), while for datasets with a hierarchical structure, \( y_k^l \) represents labels at different levels of the hierarchy. The objective is to train an encoder \( f(.) \) that learns semantically meaningful representations of the training data, \( f(x) \), which can subsequently be utilized for downstream tasks.

\subsection{Framework}
\label{subsec:framework}
Our approach is structurally similar to the standard supervised contrastive learning method \citep{khosla2020supervised}, with the key modification being the addition of multiple projection heads.
\begin{enumerate}
    \item For every sample $x_k$ where $1 \leq k \leq N$, two random augmentations $\tilde{x}_{2k}$ and $\tilde{x}_{2k-1}$ are generated, which both have the same label as $x_k$.
    
    \item The encoder network $f(.)$ takes samples $\tilde{x}_i$  and produces their respective embedding vectors where $1 \leq i \leq 2N$.
    
    \item The embedding $f(\tilde{x}_i)$ is mapped to multiple projection spaces by the projection networks $g_h$ where $h \in \{1, \dots , H\}$ and $H$ is the total number of projection heads , resulting in $z_i^h=g_h(f(\tilde{x}_i))$.
    
    \item The supervised contrastive loss is used to pull positive sample pairs together and push negative sample pairs apart in the projection spaces. Positive pairs are defined independently for each projection head based on a similarity criterion. We will elaborate on this part more in the following.
\end{enumerate}

Each projection space is dedicated to a specific similarity notion, where positive pairs are drawn closer and negative pairs are pushed apart. The similarity criteria may be based on the input samples' labels. For instance, if a pair shares the same $l$-th label, it is considered a positive pair; otherwise, it is treated as negative. Alternatively, similarity can be determined by aggregating all labels, with pairs exhibiting substantial label overlap considered positive. In this work, we focus on these two similarity notions, but other criteria could be explored in future research.

For each projection space $h$ the corresponding loss function is defined as $L_h$ inspired from \cite{khosla2020supervised}:
\begin{equation}
\label{supcon-loss}
L_h = \sum_{i \in I} \frac{-1}{|P_h(i)|} \sum_{p\in P_h(i)} \log \frac{\exp\left(\frac{z_i^h \cdot z_p^h}{\tau_h}\right)}{\sum_{a \in A(i)} \exp\left(\frac{z_i^h \cdot z_a^h}{\tau_h}\right)}.
\end{equation}

For each augmented sample, $\tilde{x}_i$, the loss function aims to increase the relative similarity between $z_i^h$ and the positive samples associated with $\tilde{x}_i$, which is defined by the indices $P_h(i)$. The similarity is measured by the dot product between the two projection vectors $z_i^h\cdot z_p^h$. We denote by $A(i)$ the set of indices for all samples, excluding sample $i$, which has a cardinality of $2N-1$. The scalar temperature hyperparameter $\tau_h$ regulates the influence of hard negatives on the final representation as discussed by \citet{chen2020simple} and its influence is further analyzed in the ablation studies.

Our final loss function is defined as $L$:
\begin{equation}
\label{mlcl-loss}
L = \sum_{h=1}^H \alpha_{h} L_{\tau_h}^h,
\end{equation}
where $\sum_{h=1}^H\alpha_{h}=1$.


We now describe each of the two tasks, multi-label classification and hierarchical classification, in more detail below.

\begin{figure}
\label{tsne-fig}
\subfloat[SupCon\label{fig:supcon_rep}]{%
  \includegraphics[height=3cm,width=.49\linewidth]{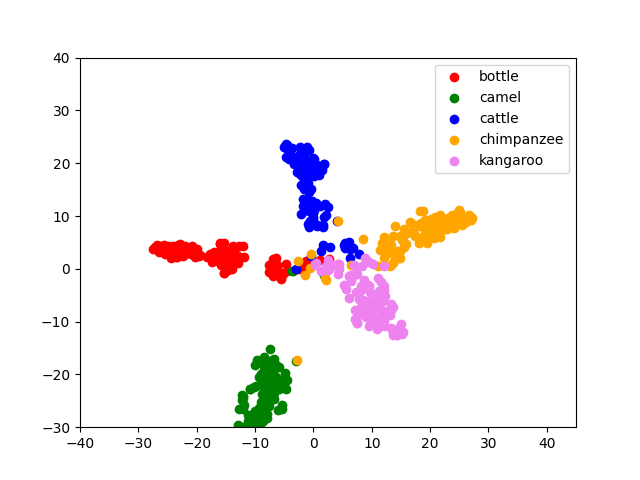}%
}
\hfill
\subfloat[MLCL\label{fig:mulhead_rep}]{%
  \includegraphics[height=3cm,width=.49\linewidth]{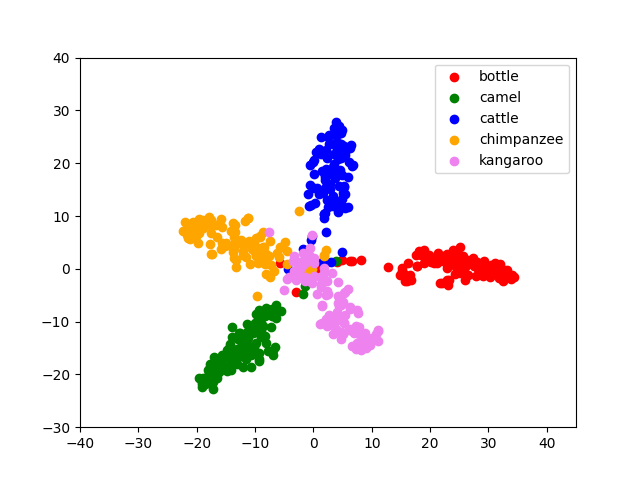}%
}
\caption{The t-SNE visualization of samples from five different classes in the representation space. The SupCon representation indicates a lack of meaningful structure, as samples from class \textit{camel} are positioned far from \textit{chimpanzee}, \textit{kangaroo}, and \textit{cattle} compared to the \textit{bottle} class. In contrast, our proposed method, MLCL, groups animal classes more closely in the representation space.}
\label{fig:representations}
\end{figure}
\subsection{Hierarchical Classification}
\label{sec:single-label}
To demonstrate this, we use the CIFAR-100 dataset \citep{cifar100} as an example. CIFAR-100 provides a two-level class hierarchy, where each sample has two labels: one for the subclass and one for the superclass. Each superclass consists of five subclasses—such as the flower superclass, which includes orchids, poppies, roses, sunflowers, and tulips. In this paper, we focus on the 100-class classification task but utilize the available superclass information to enhance the quality of the learned representations. We use multiple projection heads to deal with class hierarchies, with each head corresponding to a hierarchy level. Taking the CIFAR-100 dataset as an example with two hierarchy levels, our framework utilizes two projection heads: one for the standard subclass labels and one for the superclasses. Figure \ref{fig:img-arch} shows the architecture with the two projection heads. 

During training, given an input batch, we apply the first augmentation module to create a duplicate batch. Both batches pass through an encoder to obtain normalized embeddings. In the training stage, these embeddings are given to the two projection heads. The first head pulls samples with the same subclass label together using the loss function defined in Equation \eqref{supcon-loss}, while the second head pulls the samples with the same superclass together. We set a higher temperature for the second head to achieve a more distinct representation for samples belonging to different superclasses. According to \citet{wang2021understanding}, increasing the temperature leads to greater separation among the distinct classes, while decreasing the temperature results in a more evenly distributed set of classes, making the model more tolerant to negative samples.
In the inference stage, these two heads are removed, leaving only a single representation that has integrated information from both similarity criteria. For evaluation, a linear classifier is trained on these frozen representations using a cross-entropy loss, which is also known as the linear evaluation protocol in representation learning~\citep{chen2020simple, khosla2020supervised}.

\begin{table*}[t]
\caption{Overview of evaluation datasets: 
CIFAR-100 and DeepFashion each have two hierarchical levels. However, classification within the subclass level is the primary focus, while the superclass serves as an auxiliary signal during training.}
\label{overview}
\vskip 0.15in
\hspace{-1cm} 
\begin{center}
\begin{small}
\begin{sc}
\begin{tabular}{cccccccc}
\toprule
Dataset & Task  & \#Train & \#Test & \#Levels & \#Classes \\
\midrule
CIFAR-100    & image classification& 50K& 10K& 2 & 100 \\
DeepFashion  & image classification & 200K & 40K & 2 & 50 \\
TripAdvisor    & sentiment analysis& 10K & 2K & 7 & 3 \\
BeerAdvocate & sentiment analysis& 10K & 2K & 5 & 3 \\
\bottomrule
\end{tabular}
\end{sc}
\end{small}
\end{center}
\vskip 0.15in
\end{table*}
\subsection{Multi-label Classification}
\label{sec:multi-label}
Multi-label classification aims to predict $L$ labels for each sample $x_k$, where each sample has a multi-hot label vector $y_k\in \{0,1\}^L$ and $y_k^l$ denotes the one-hot representation of the $l$-th label. Similarly to the method described in Section~\ref{sec:single-label}, we assign a distinct projection head for each label and include an additional global projection head to capture broader semantic relationships which will be elaborated on later.
We demonstrate our method for multi-label classification using the TripAdvisor review dataset \citep{wang2010latent}. Each review has $L=7$ labels that indicate user ratings for different aspects such as service, location, etc. We employ $H=8$ total heads, where seven heads are dedicated to the seven different aspects, and the final eighth head (global head) is used to focus on global similarities between two hotel reviews that have a similar overall sentiment, combining all aspects together. This illustrates another application of projection heads, where they are used as regularizers, especially in scenarios with limited training samples. Figure \ref{fig:text-arch} shows the architecture of the model.

\paragraph{Global Projection Head}
For datasets with ordinal class numbers, we can define a global similarity perspective that combines the information from all classes. For instance, in the hotel review dataset, if two reviews have 4 or 5 ratings for all aspects, then they are generally similar, and we expect them to have a similar representation, even if the specific rating varies for different aspects, such as one review giving four to service and the other giving five. We use the Jaccard similarity metric, inspired by \citet{zaigrajewcontrastive} to focus on these global similarities. 

The Jaccard similarity measure $s_{i,j}$ is defined for two samples $i$ and $j$ as: 
$$s_{i, j} = \frac{\sum_{l=1}^{L} \min(y_{i}^l, y_{j}^l)}{\sum_{l=1}^{L} \max(y_{i}^l, y_{j}^l)}$$

We introduce a hyperparameter $t$ to define all samples $i$ and $j$ with $s_{i,j} > t$ as positive samples. The loss function defined in Equation \eqref{supcon-loss} is then adjusted to put more weight on samples with higher Jaccard similarity as proposed by \citet{zaigrajewcontrastive}: 
\begin{equation*}
L_{gl} = \sum_{i \in I} \frac{-1}{|P(i)|} \sum_{p\in P(i)} s_{i, p}\log \frac{\exp\left(\frac{z_i \cdot z_p}{\tau}\right)}{\sum_{a \in A(i)} \exp\left(\frac{z_i \cdot z_a}{\tau}\right)}.
\end{equation*}

The training procedure for multi-label classification closely follows the approach used for single-label classification in Section \ref{sec:single-label}. However, since our target datasets are textual, we follow the common practice of combining contrastive learning with the cross-entropy loss, as demonstrated by \citet{gunel2020supervised}. The remainder of the training process is consistent with Section~\ref{sec:single-label}.

The linear classifier and cross-entropy loss are trained simultaneously with the projection heads and the pretrained Bert encoder. The final loss $L_{\text{MLCL}}$ is  defined as a weighted average of the loss in each projection head plus the cross-entropy loss, denoted as 
\begin{equation}
\label{multi-label-loss}
L = \sum_{h=1}^{H} \alpha_{h} L^h_{\tau_h} + (1-\sum_{h=1}^{H}\alpha_h) L_{\text{ce}},
\end{equation}
where $L_{ce}$ denotes the cross-entropy loss used for linear classification.

\subsection{Temperature Analysis}
\label{temperature-analysis}
Temperature plays a crucial role in controlling the impact of hard negative samples during training. Hard negatives are particularly challenging to distinguish from positive samples, as they often have high similarity to the anchor in the representation space. The temperature parameter helps to adjust the sensitivity of the model to subtle differences, allowing it to focus more effectively on more difficult distinctions. As temperature decreases, the model concentrates more on a smaller set of the closest hard negative samples, applying a stronger penalty to these examples. In contrast, when the temperature is large, the model distributes the penalty more evenly across all negative samples. To demonstrate this, we analyze two extreme cases of $\tau \to 0^+$  and $\tau \to +\infty$, inspired by \cite{wang2021understanding}. Based on Equation \eqref{supcon-loss}, the loss function for a given sample $i$ in the projection space $h$ is denoted as $L_h^i$:
\begin{equation*}  
L^i_h = \frac{-1}{|P_h(i)|} \sum_{p\in P_h(i)} \log \frac{\exp\left(\frac{z_i^h \cdot z_p^h}{\tau_h}\right)}{\sum_{a \in A(i)} \exp\left(\frac{z_i^h \cdot z_a^h}{\tau_h}\right)}.
\end{equation*}

For the sake of better readability, the index $h$ is omitted in the following derivation:
\begin{equation*}
\begin{split}
&\lim_{\tau \to 0^+} |P(i)| \times L^i = 
\\
&\sum_{p\in P(i)} \lim_{\tau \to 0^+} \Bigg[-\log \frac{\exp(z_i \cdot z_p/\tau)}{\sum_{a \in A(i)} \exp(z_i \cdot z_a/\tau)}\Bigg] = 
\\
&\sum_{p\in P(i)} \lim_{\tau \to 0^+} \Bigg[-\log \frac{\exp\big(z_i \cdot (z_p-z_i^{\max})/\tau\big)}{\sum_{a \in A(i)} \exp\big(z_i \cdot (z_a-z_i^{\max})/\tau\big)}\Bigg]
 \\
  &=\sum_{p\in P(i)} \lim_{\tau \to 0^+} \Bigg[ - \big(zi \cdot (z_p - z_i^{\max})/\tau\big) 
+ \\
 &  \hspace{34pt} \log \bigg(1 + \sum_{a \in A(i) \backslash p }\exp \big(z_i \cdot (z_a - z_i^{\max})/\tau\big)\bigg) \Bigg]=
  \\
  &\sum_{p\in P(i)} \lim_{\tau \to 0^+} \Bigg[
   - \big(z_i \cdot (z_p - z_i^{\max})/\tau\big)\Bigg] =
  \\
  & = \lim_{\tau \to 0^+} \frac{1}{\tau} \bigg[|P(i)| z_i \cdot z_i^{\max} - \sum_{p\in P(i)} z_i \cdot z_p\bigg],
\end{split}
\end{equation*}
where $z_i^{\max}=\arg\max_{a \in A} z_i \cdot z_a$.  We can infer that as \( \tau \to 0^+ \), the loss function focuses primarily on the positive samples and the negative sample with the maximum similarity to $z_i$, meaning only the hardest negative is considered. In contrast, as $\tau$ increases, negative samples contribute more uniformly to the loss function (for the derivation as $\tau \to +\infty$ please refer to the appendix). In hierarchical classification, the primary task is to classify at the lowest level (subclass), while higher-level labels in the hierarchy (superclass) are used to create a more structured feature space. To achieve this, we use a lower temperature for the projection space corresponding to the lowest level of the hierarchy to fully separate different subclasses, while using a higher temperature for superclasses to uniformly learn from all negative samples. Gradient analysis in the appendix demonstrates a similar effect for hard positives. At low temperatures, the loss function concentrates on the positive sample with the lowest similarity to the anchor, which is not ideal for higher levels of the hierarchy. Two positive samples at the superclass level may belong to different subclasses, requiring some degree of separation. Hence, focusing solely on hard positives at higher hierarchical levels can be counterproductive. This provides a second rationale for using a lower temperature for subclasses to ensure complete separation and a higher temperature for superclasses to promote more uniform learning across all samples. Further empirical analysis is provided in Section~\ref{experiments}.

\setlength{\tabcolsep}{6pt} 
\begin{table*}[t]
\caption{Top-1 classification accuracy on CIFAR-100 and DeepFashion test sets. We compare MLCL with ResNet50 trained with cross-entropy (CE), SimCLR~\citep{chen2020simple}, SupCon~\citep{khosla2020supervised}, Guided~\citep{landrieu2021leveraging}, and HiMulConE~\citep{zhang2022use} losses. The highest test accuracy is marked in bold.}
\label{deepfashion-cifar100-results}
\vskip 0.15in
\begin{center}
\begin{small}
\begin{sc}
\begin{tabular}{cccccccc}
\toprule
Dataset & SimCLR & CE & SupCon & Guided & HiMulConE & MLCL (ours) \\
\midrule
Cifar-100 & 70.70 & 75.30 & 76.50 & 76.40 & - & \textbf{77.70} \\
DeepFashion & 70.38 &  72.44 & 72.82 & 72.61 & 73.21 & \textbf{73.90} \\
\bottomrule
\end{tabular}
\end{sc}
\end{small}
\end{center}
\vskip -0.1in
\end{table*}

\begin{table*}[ht]
\caption{Top-1 classification accuracy on the test set of CIFAR-100 dataset using a subset of training samples. The highest test accuracy is marked in bold.}
\label{cifar100-results}
\vskip 0.15in
\begin{center}
\begin{small}
\begin{sc}
\begin{tabular}{rrrrr}
\toprule
\#Training Samples & SimCLR & Cross-Entropy & SupCon & MLCL (ours) \\
\midrule
1,000 & 19.31 & 20.50 & 26.82 & \textbf{34.74} \\
5,000 & 34.51 &  21.56 & 46.15 & \textbf{56.02} \\
10,000 & 40.20 & 42.25 & 49.87 & \textbf{59.32} \\
20,000 & 54.73 &  59.04 & 65.21 & \textbf{69.36} \\
30,000 & 59.26 &  63.47 & 71.49 & \textbf{72.50} \\
40,000 & 63.21 &  67.10 & 74.80 & \textbf{75.60} \\
\bottomrule
\end{tabular}
\end{sc}
\end{small}
\end{center}
\vskip -0.1in
\end{table*}

\section{Experiments}
\label{experiments}
We evaluate the presented method using four benchmark datasets, two from the image domain and two text datasets. The first task is image classification on CIFAR-100 \citep{cifar100}, a widely used benchmark for image classification, and DeepFashion \citep{liuLQWTcvpr16DeepFashion}, a benchmark for hierarchical classification. We selected CIFAR-100 and DeepFashion because they offer class hierarchies. The second application focuses on multi-label classification using two text datasets for aspect-based sentiment analysis: TripAdvisor \citep{wang2010latent} and BeerAdvocate \citep{mcauley2012learning}.

\subsection{Datasets}
CIFAR-100 comprises 50,000 training images and 10,000 testing images, categorized into 100 classes, which are further grouped into 20 superclasses. The DeepFashion \citep{liuLQWTcvpr16DeepFashion} dataset, a large-scale clothing dataset, contains 200,000 training images and 40,000 testing images, with labels spanning 50 categories and organized into three superclasses. We intentionally chose these datasets because the widely used ImageNet \citep{deng2009imagenet} lacks groundtruth superclasses. This limitation makes ImageNet less appropriate for our multi-level setting, as it would reduce the problem to the standard supervised contrastive approach (SupCon). For multi-label classification, we conduct experiments on two textual datasets, TripAdvisor and BeerAdvocate. TripAdvisor is a hotel review dataset, with each review having seven different ratings based on various aspects (value, room, location, cleanliness, service, overall). We map the ratings to sentiments by changing ratings four and five to positive, three to neutral, and the others to negative. Similarly, BeerAdvocate consists of five aspects (appearance, aroma, palate, taste, overall), each associated with three sentiments, mirroring the structure of the other dataset. The summary of the datasets is given in Table \ref{overview}.
\setlength{\tabcolsep}{2pt} 
\begin{table*}[ht]
\caption{Top-1 classification accuracy on TripAdvisor and BeerAdvocate datasets. A direct comparison with SupCon \citep{khosla2020supervised} is not feasible due to its single-label classification nature. Instead, we compare our results with fine-tuned BERT using cross-entropy (CE) loss and our proposed MLCL loss (without and with the global projection head). The highest accuracy is highlighted in bold.}
\label{multi-label-results}
\vskip 0.15in
\begin{center}
\begin{small}
\begin{sc}
\begin{tabular}{ccccc}
\toprule
Dataset & \#Samples & CE & MLCL w/o $h_{global}$ & MLCL (ours) \\
\midrule
\multirow{4}{*}{TripAdvisor} & 30  & 71.30 &  72.50 & \textbf{72.9} \\
 & 60 & 74.19 &  74.81 & \textbf{75.1} \\
 & 180  & 77.00 &  77.30 & \textbf{77.9} \\
 & 360 & 78.10 &  78.44 & \textbf{79.0} \\
\midrule
\multirow{4}{*}{BeerAdvocate} & 30 & 65.45 & 65.95 & \textbf{66.90} \\
 & 60 & 66.37 &  66.71 & \textbf{67.40} \\
 & 180 & 68.70 &  69.25 & \textbf{69.61} \\
 & 360 & 70.54 &  71.22 & \textbf{71.81} \\
\bottomrule
\end{tabular}
\end{sc}
\end{small}
\end{center}
\vskip -0.1in
\end{table*}
\subsection{Implementation Details}
\paragraph{Hierarchical Classification} We conduct experiments on the CIFAR-100 and DeepFashion datasets with $H=2$ projection heads, assigning the first head to the subclass and the second to the superclass labels. The parameters for the first head are set as $\tau_1=0.1$ and $\alpha_1=0.5$, and for the second head as $\tau_2=0.5$, $\alpha_2=0.5$, following the notation in Equation \eqref{mlcl-loss}. The remaining hyperparameters are set as proposed in the SupCon paper \citep{khosla2020supervised} to ensure a fair comparison with this established baseline. Our model architecture includes a ResNet-50 encoder \citep{he2016deep} and two multi-layer perceptions (MLPs) with a single hidden layer serving as projection heads. The model is trained for 250 epochs, which is a quarter of the 1,000 epochs used by SupCon, with a batch size of 512. We use stochastic gradient descent (SGD) with momentum optimizer~\citep{ruder2016overview}. After training, the projection heads are discarded, and a linear classifier is trained on the frozen learned representation from the encoder's output to obtain the final accuracy on the subclass labels. This standard evaluation method, known as the linear evaluation protocol, is commonly used to assess the representations learned in contrastive learning studies \citep{khosla2020supervised, chen2020simple}. The procedure is depicted in Figure \ref{fig:img-arch}.

\paragraph{Multi-label Classification} We evaluate our approach on two text datasets for multi-label classification: TripAdvisor hotel reviews and BeerAdvocate reviews. The former consists of seven labels per review, and the latter has five labels. The task is to predict the sentiment for each aspect, a multi-label classification problem. We employ a pretrained BERT encoder \citep{devlin2018bert} with 512 embedding dimensions and fine-tune it by back-propagating the loss from multiple projection heads and a fully connected layer linked to the output of the encoder. The framework of our approach is illustrated in Figure \ref{fig:text-arch}. For the TripAdvisor and BeerAdvocate datasets, we use $H=8$ and $H=6$ projection heads, respectively. Each projection head corresponds to one label, with an additional global projection head for incorporating global similarities into the learned representation. The cross-entropy loss contribution is maintained at $0.7$ for both datasets. For the TripAdvisor dataset, we set $\alpha_i=0.03$ for the first seven heads and $\alpha_8=0.1$ for the final head. For the BeerAdvocate dataset, we apply $\alpha_i=0.04$ for the first five heads and $\alpha_6=0.1$ for the global head. The model is fine-tuned for 100 epochs using the Adam optimizer \citep{kingma2014adam} with a learning rate of $1e-5$ and a batch size of 16. We found that larger batch sizes did not significantly improve performance.

\subsection{Results}

\paragraph{Hierarchical Classification} 
Our method demonstrates a marginal improvement of 1\% over SupCon when trained on the complete CIFAR-100 and DeepFashion training datasets, as presented in Table \ref{deepfashion-cifar100-results}. Notably, this performance is achieved significantly faster, requiring only 250 training epochs compared to SupCon's 1,000 epochs. Furthermore, our approach exhibits superior performance with smaller training sets, achieving an enhancement of 9\% to 10\% for dataset sizes of 10K and 5K samples, as detailed in Table \ref{cifar100-results}. Additionally, our method surpasses the performance of models trained exclusively with cross-entropy loss by a substantial margin.

\paragraph{Multi-label Classification} Our results on the TripAdvisor and BeerAdvocate datasets are reported in Table~\ref{multi-label-results}. The numbers represent the average accuracies across ten different seeds. Using a pretrained BERT encoder, we focus on scenarios with limited training samples ranging from 30 to 360 while maintaining a test dataset size of 2K for all experiments. Our findings indicate that our approach enhances the quality of the learned representation, even with a robust pretrained encoder. The positive impact of the global projection head is also evident when comparing the third and fourth columns in Table~\ref{multi-label-results}.

\subsection{Ablation Study}
\label{ablation}
\begin{figure}[ht] 
  \centering
  \includegraphics[width=\linewidth]{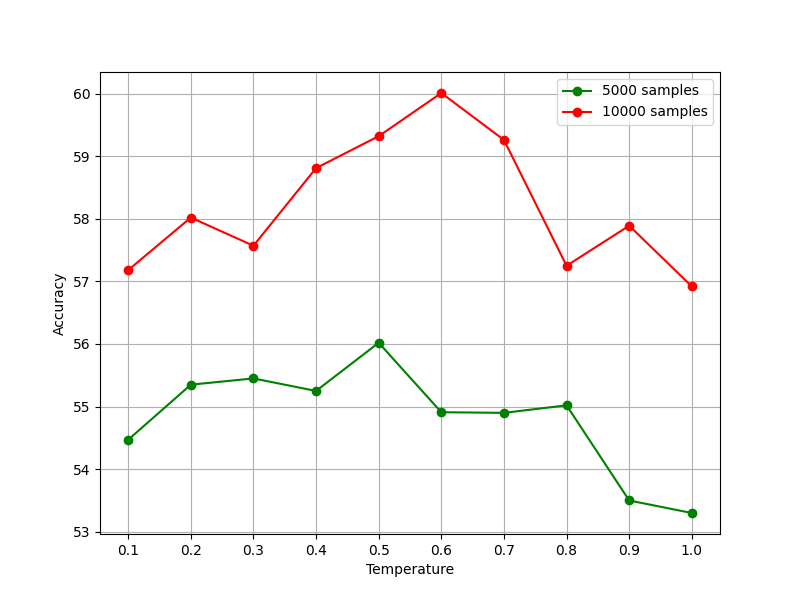}
  \caption{Accuracy of MLCL as a function of the superclass projection head temperature. Increasing the temperature to 0.5 reduces the network's focus on hard negatives, leading to improved accuracy.}
  \label{fig:temp-analysis}
\end{figure}

The feature spaces of our approach and SupCon are visualized in Figure \ref{fig:representations} using t-SNE \citep{van2008visualizing}. The second projection head helps maintain the proximity of representations for samples within the same superclass. The class \textit{camel} is expected to be closer to \textit{cattle}, \textit{chimpanzee}, and \textit{kangaroo}, while remaining distant from \textit{bottle}, which belongs to a different superclass. Furthermore, Figure \ref{fig:temp-analysis} illustrates the impact of the superclass projection head's temperature on the final accuracy using a subset of training samples. Both excessively high and low temperatures negatively affect performance, while $\tau = 0.5$ yields the best results. The subclass temperature is fixed at $0.1$. For additional ablation studies on performance under label noise and transfer learning, please refer to the appendix.

\section{Conclusion}
\label{conclusion}

We introduced a novel supervised contrastive learning method within a unified framework called multi-level contrastive learning, which generalizes to tasks such as hierarchical and multi-label classification. Our approach incorporates multiple projection heads into the encoder network to learn representations at different levels of the hierarchy or for each label. This allows the method to capture label-specific similarities, exploit class hierarchies, and act as a regularizer to prevent overfitting. Although the parameter size for MLCL is equivalent to that of SupCon, it converges more rapidly and learns more effective representations, especially when training samples are limited. We have analyzed the impact of temperature on each projection head and showed that a higher superclass temperature improves the representations by allowing the model to focus less on hard negative samples. As shown in Section \ref{experiments}, this approach improves representation quality across various settings and datasets. Notably, the approach integrates seamlessly with existing contrastive learning methods without relying on any specific network architecture, making it flexible and applicable to a wide range of downstream tasks. Future work could explore the interpretability and explainability of the learned representations.

\bibliography{sample_paper}

\clearpage
\onecolumn
\aistatstitle{Supplementary Materials}

\section{Temperature Analysis for $\tau \to +\infty$}
We showed that adjusting the temperature parameter in the contrastive loss influences the focus of the model on different samples. In the main paper, we derived the case where $\tau \to 0^+$, illustrating that this leads the model to concentrate exclusively on the hardest negative samples among all available negatives. Here, we extend the analysis to the case where $\tau \to +\infty$. The loss function for a given sample 
$i$ in the projection space $h$ is denoted as:
\begin{equation*}  
L^i_h = \frac{-1}{|P_h(i)|} \sum_{p\in P_h(i)} \log \frac{\exp\left(\frac{z_i^h \cdot z_p^h}{\tau_h}\right)}{\sum_{a \in A(i)} \exp\left(\frac{z_i^h \cdot z_a^h}{\tau_h}\right)}.
\end{equation*}
As before, we omit the index $h$ for easier readability:

\begin{align*}
\lim_{\tau \to +\infty} |P(i)| \times L^i &= \sum_{p\in P(i)} \lim_{\tau \to +\infty} \Bigg[-\log \frac{\exp(z_i \cdot z_p/\tau)}{\sum_{a \in A(i)} \exp(z_i \cdot z_a/\tau)}\Bigg] \\
&= \sum_{p\in P(i)} \lim_{\tau \to +\infty} \Bigg[-z_i \cdot z_p /\tau + \log\sum_{a\in A(i)}\exp(z_i \cdot z_a/\tau)\Bigg] \\
&= \sum_{p\in P(i)} \lim_{\tau \to +\infty} \Bigg[-z_i \cdot z_p /\tau + \log(2N-1) + \sum_{a \in A(i)}\frac{\exp(z_i \cdot z_a/\tau)-(2N-1)}{2N-1}\Bigg]\\
&= \sum_{p\in P(i)} \lim_{\tau \to +\infty} \Bigg[-z_i \cdot z_p /\tau + \log(2N-1) -1 + \sum_{a \in A(i)}\frac{\exp(z_i \cdot z_a/\tau)}{2N-1}\Bigg]\\
&= \sum_{p\in P(i)} \lim_{\tau \to +\infty} \Bigg[-z_i \cdot z_p /\tau + \log(2N-1) -1 + \sum_{a \in A(i)}\frac{1 + z_i \cdot z_a/\tau}{2N-1}\Bigg]\\
&= \sum_{p\in P(i)} \lim_{\tau \to +\infty} \Bigg[-z_i \cdot z_p /\tau + \log(2N-1) + \frac{1}{\tau}\sum_{a \in A(i)}\frac{z_i \cdot z_a}{2N-1}\Bigg]\\
&= \lim_{\tau \to +\infty} \Bigg[|P(i)|\log(2N-1) + \frac{1}{\tau}|P(i)|\sum_{a \in A(i)}\frac{z_i \cdot z_a}{2N-1} - \sum_{p\in P(i)} z_i \cdot z_p /\tau \Bigg]
\end{align*}
The third equality results from the Taylor expansion of $\log(\sum_{a\in A(i)}\exp (z_i\cdot z_a/\tau)$ as $\tau \to +\infty$, while the fifth equality follows from the Taylor expansion of $\exp (z_i\cdot z_a/\tau)$ in the same limit.\newline
We observe that as $\tau \to +\infty$, the loss treats all the negative samples equally, meaning their contribution to the loss becomes uniform. Using a higher temperature for the superclass projection head encourages more uniform learning across all negative samples, allowing for greater flexibility in capturing dissimilarities.

\section{Gradient Analysis}
For clarity, we omit the index $h$ in our gradient computation. The final loss function $L$ is the sum of the losses for each individual sample, $L = \sum_{t \in I} L_i$. Unlike SupCon which only provides the derivative of $L^i$ with respect to $z_i$, we compute the derivative for all individual loss terms $L^t$ where $t \in I$.

 We calculate the derivative of $L^t$ by dividing it to three different cases: 1. $t=i$,  2. $t \in P(i)$ and 3. $t \in A(i) \setminus P(i)$. The final gradient of $L$ with respect to $z_i$ is then obtained by summing these individual gradients. The relative similarity between samples $m$ and $n$ is defined as $S_{m,n}$, where

\[
S_{m,n} = \frac{\exp(z_m \cdot z_n / \tau)}{\sum_{a \in A(m)} \exp(z_m \cdot z_a / \tau)}
\]

For the first case, where $t=i$, the gradient is:

\begin{align*}
\frac{\partial {L^t}}{\partial z_i} &= \frac{\partial}{\partial z_i} \Bigg[\frac{-1}{|P(i)|} \sum_{p \in P(i)}\log 
\frac{\exp( z_i \cdot z_p / \tau )}{\sum_{a \in A(i)} \exp( z_i \cdot z_a / \tau )}\Bigg]
\\
&=\frac{-1}{|P(i)|}\sum_{p \in P(i)}\frac{\partial}{\partial z_i}\Bigg[ \frac{z_i \cdot z_p}{\tau} - \log\bigg[\sum_{a \in A(i)} \exp( {z_i \cdot z_a}/{\tau} )\bigg] \Bigg]
=\frac{-1}{|P(i)|\tau}\sum_{p \in P(i)}  \bigg[z_p  - \frac{\sum_{a \in A(i)} z_a \exp ( z_i \cdot z_a / \tau )}{\sum_{a \in A(i)} \exp ( z_i \cdot z_a / \tau )}\bigg]  \\
&= \frac{-1}{\tau |P(i)|} \sum_{p \in P(i)} \bigg[ z_p - \sum_{a \in A(i)} z_aS_{i,a}\bigg] = \frac{-1}{\tau |P(i)|}
\sum_{p \in P(i)} z_p + \frac{1}{\tau}\sum_{a \in A(i)} z_aS_{i,a}
\end{align*}

The second case where $t \in P(i)$:

\begin{align*}
\frac{\partial {L^t}}{\partial z_i} &= \frac{\partial}{\partial z_i} \Bigg[\frac{-1}{|P(t)|} \sum_{p \in P(t)}\log 
\frac{\exp( z_t \cdot z_p / \tau )}{\sum_{a \in A(t)} \exp( z_t \cdot z_a / \tau )}\Bigg]
\\
&=\frac{-1}{|P(t)|}\sum_{p \in P(t)}\frac{\partial}{\partial z_i}\Bigg[ \frac{z_t \cdot z_p}{\tau} - \log\bigg[\sum_{a \in A(t)} \exp( {z_t \cdot z_a}/{\tau} )\bigg] \Bigg]
=\frac{-1}{|P(i)|\tau}\bigg[z_t -\sum_{p \in P(t)} \frac{ z_t \exp ( z_t \cdot z_i / \tau )}{\sum_{a \in A(t)} \exp ( z_t \cdot z_a / \tau )}\bigg]  \\
&= \frac{-1}{\tau |P(t)|} \bigg[ z_t - |P(t)|z_tS_{t,i}\bigg] = \frac{-1}{\tau |P(i)|} z_t + \frac{1}{\tau} z_tS_{t,i}
\end{align*}

where we used the fact that $|P(i)| = |P(t)|$. Similarly for the third case where $t \in A(i) \setminus P(i)$:

\begin{align*}
\frac{\partial {L^t}}{\partial z_i} &= \frac{1}{\tau} z_tS_{t,i}
\end{align*}

Thus, the gradient of the loss function with respect to $z_i$ is:

\begin{align*}
\frac{\partial {L}}{\partial z_i} &=
\frac{\partial}{\partial z_i} L_i + \sum_{j \in P(i)} \frac{\partial}{\partial z_i} L_j + \sum_{n \in A(i) / P(i)} \frac{\partial}{\partial z_i} L_n 
\\
&= \frac{-1}{\tau |P(i)|}
\sum_{p \in P(i)} z_p + \frac{1}{\tau}\sum_{a \in A(i)} z_aS_{i,a} + \sum_{j \in P(i)} \Big( \frac{-1}{\tau |P(i)|} z_j + \frac{1}{\tau} z_jS_{j,i} \Big) + \sum_{n \in A(i) / P(i)}\frac{1}{\tau} z_nS_{n,i}
\\
&= -\frac{2}{\tau} \Bigg[ 
\frac{1}{|P(i)|}\sum_{p \in P(i)} z_p - \sum_{a \in A(i)} z_a\frac{S_{a,i}+S_{i,a}}{2} 
\Bigg]
\end{align*}
In each iteration, $z_i$ is updated by moving in the direction of the negative gradient, which adjusts $z_i$ towards the average of the positive samples, offset by the weighted average of all samples. This aligns with the intuition that each sample seeks to have a representation closer to its class while distinct from others. The effect of temperature is evident in the relative similarities $S_{i,a}$ and $S_{a,i}$. A lower temperature leads to a sharp $S_{i,a}$ (and $S_{a,i}$), which is mostly dependent on the direction of the sample with maximum similarity $z_i^{\max}=\arg\max_{a \in A(i)} z_i \cdot z_a$. In this case, easy positive pairs—those with large values of $S_{p,i}$ or $S_{i,p}$—will have their influence either canceled out or reversed by significant negative contributions in the second term of the derivative. Consequently, the model will primarily focus on bringing the representations of hard positive pairs closer together, which is undesirable at higher levels of hierarchical classification.  For example, at the second level of the CIFAR-100 classification, a hard positive pair could be trout and shark, which are distinct classes within the superclass 'fish', and some degree of separability between their representations is desired. This observation further supports the use of a higher temperature parameter for the projection spaces corresponding to higher levels of the hierarchy.

\section{Ablation Studies}
\label{ablation}
\subsection{Label Noise} We evaluate the impact of the global projection head in our approach by introducing uniform label noise into the TripAdvisor and BeerAdvocate training datasets. For a label with a total of $N$ classes, the noise level of $r\%$ implies that the label $y_k$ for sample $x_k$ remains $y_k$ with a probability of $1-\frac{r(N-1)}{N}$ after adding noise, and switches to any of the other $(N-1)$ labels with a probability of $\frac{r}{N}$. This results in a completely random label if $r=100\%$. We train the model with label noise rates of $30\%$, $50\%$, $70\%$, and $80\%$. The corresponding results are presented in Table \ref{noise}. Our observation indicates that models trained with a large weight for global projection loss, set at $0.6$ for this experiment, show increased robustness to label noise, which is consistent with our interpretation of the global head as a regularizer.
\begin{table*}[h]
\caption{The Top-1 classification accuracy is reported under uniform noise labels for the TripAdvisor and BeerAdvocate datasets. In this setup, the global projection head is assigned a weight of 
$0.6$, while the total weight for the other projection heads remains constant at 
$0.2$, resulting in a cross-entropy weight of 
$0.2$. Results indicate an improvement of up to 7\% in noisy label settings when assigning a large weight to the global projection head loss.}
\label{noise}
\vskip 0.15in
\begin{center}
\begin{small}
\begin{sc}
\begin{tabular}{cccccccccc}
\toprule
\textbf{Dataset} & \textbf{\#Samples} & \multicolumn{4}{c}{\textbf{Cross-Entropy}} & \multicolumn{4}{c}{\textbf{MSCL (ours)}} \\
\cmidrule(lr){3-6} \cmidrule(lr){7-10}
& & 30\% & 50\% & 70\% & 80\% & 30\% & 50\% & 70\% & 80\%\\
\midrule
\multirow{4}{*}{TripAdvisor} & 60 & 61.00 & 51.77 & 40.4 & 36.00 & \textbf{62.75} & \textbf{55.00} & \textbf{43.11} & \textbf{37.00}\\
 & 180 & 61.17 & 49.83 & 39.74 & 36.31 & \textbf{64.10} & \textbf{54.79} & \textbf{44.30} & \textbf{39.10}\\
 & 360 & 62.28 & 50.96 & 40.25 & 36.54 & \textbf{64.90} & \textbf{55.43} & \textbf{43.72} & \textbf{39.59}\\
 & 600 & 63.00 & 50.72 & 40.89 & 37.13 & \textbf{65.00}& \textbf{55.50} & \textbf{44.90} & \textbf{40.24}\\
\midrule
\multirow{4}{*}{BeerAdvocate} & 60 & 57.59 & 45.53 & 36.54 & 34.3 & \textbf{62.18} & \textbf{53.70} & \textbf{39.57} & \textbf{36.47} \\
 & 180 & 60.27 & 47.23 & 38.46 & 35.90 & \textbf{63.67} & \textbf{53.81} & \textbf{42.88} & \textbf{37.98}\\
 & 360 & 58.20 & 47.49 & 39.66 & 36.79 & \textbf{63.63} & \textbf{54.83} & \textbf{44.72} & \textbf{41.05}\\
 & 600 & 60.90 & 49.00 & 40.61 & 36.40 &  \textbf{64.78} & \textbf{55.86} & \textbf{43.30} & \textbf{39.58}\\
\bottomrule
\end{tabular}
\end{sc}
\end{small}
\end{center}
\vskip -0.1in
\end{table*}
\subsection{Transfer Learning}
To demonstrate the generalizability of our model's learned features, we conducted a transfer learning experiment. First, we train the encoder on CIFAR-100 and then train a linear classifier on CIFAR-10 using the frozen encoder weights from CIFAR-100. This approach enables us to assess the transferability of features learned on CIFAR-100 to CIFAR-10. The results, presented in Table \ref{results-transferlearning}, compare the performance of SupCon and our method, demonstrating an increase of over 1\% across various training sample sizes.

\begin{table}[ht]
\caption{Features from CIFAR-100 transferred to CIFAR-10. Classification accuracy of a linear classifier trained on a frozen encoder from the CIFAR-100 dataset. The highest test accuracy is highlighted in bold.}
\label{results-transferlearning}
\vskip 0.15in
\begin{center}
\begin{small}
\begin{sc}
\begin{tabular}{ccc}
\toprule
\#Training Samples & SupCon & MSCL (ours) \\
\midrule
50K & 85.97 &  \textbf{86.88}  \\
40K & 83.70 & \textbf{84.91}  \\
30K & 82.54 & \textbf{83.69}  \\
20K & 78.62 & \textbf{79.44}  \\

\bottomrule
\end{tabular}
\end{sc}
\end{small}
\end{center}
\vskip -0.1in
\end{table}

\newpage
\subsection{Feature Analysis}
\begin{figure}[h]
\label{tsne-fig2}
\subfloat[SupCon\label{fig:supcon_rep2}]{%
  \includegraphics[height=5cm,width=.49\linewidth]{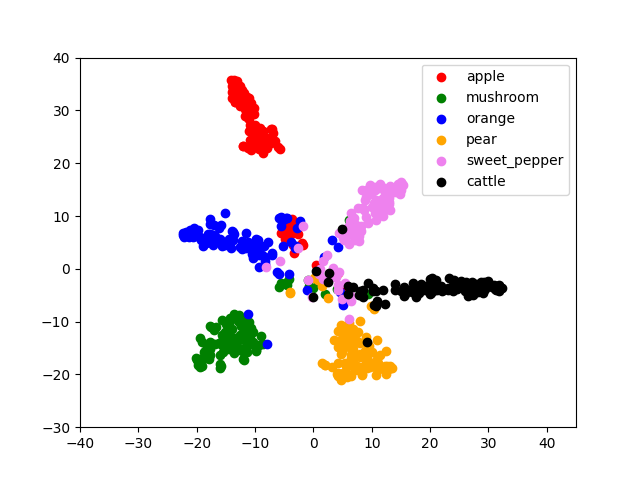}%
}
\hfill
\subfloat[MLCL\label{fig:mulhead_rep2}]{%
  \includegraphics[height=5cm,width=.49\linewidth]{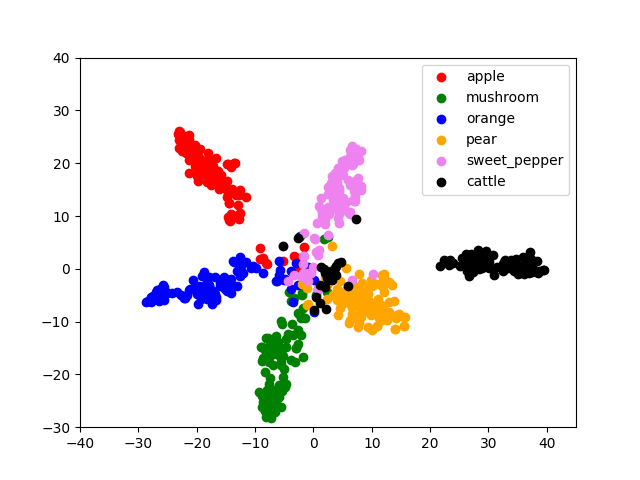}%
}
\caption{SupCon positions samples from the \textit{cattle} class close to \textit{pear} and \textit{sweet pepper}. In contrast, MLCL clusters the fruit and vegetable classes more closely in the representation space while separating them from the \textit{cattle} class.}
\label{fig:representations2}
\end{figure}

\begin{figure}
\label{tsne-fig3}
\subfloat[SupCon\label{fig:supcon_rep}]{%
  \includegraphics[height=5cm,width=.49\linewidth]{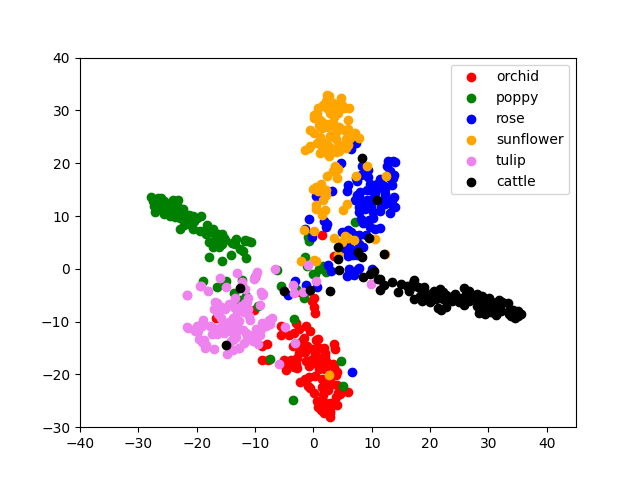}%
}
\hfill
\subfloat[MLCL\label{fig:mulhead_rep}]{%
  \includegraphics[height=5cm,width=.49\linewidth]{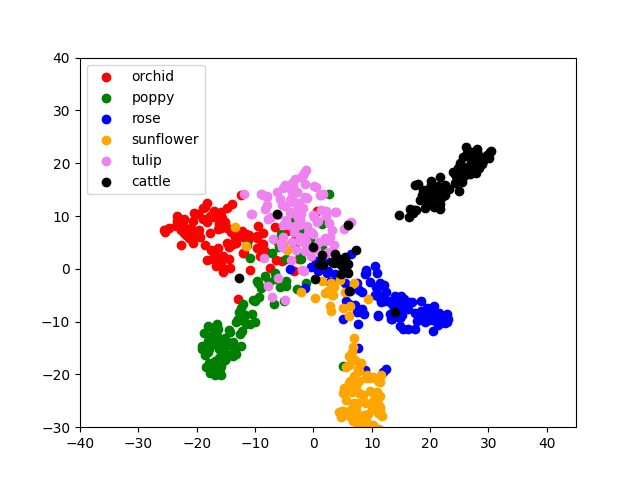}%
}
\caption{MLCL groups the flower classes more tightly in the representation space while separating them from the class \textit{cattle}.}
\label{fig:representations3}
\end{figure}

\begin{figure}
\label{tsne-fig4}
\subfloat[SupCon\label{fig:supcon_rep}]{%
  \includegraphics[height=5cm,width=.49\linewidth]{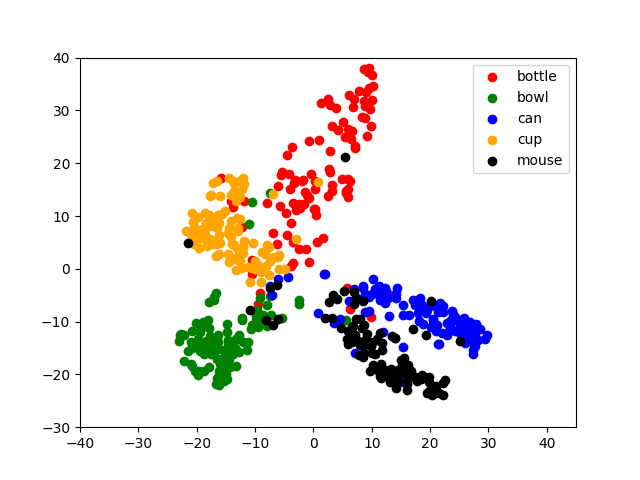}%
}
\hfill
\subfloat[MLCL\label{fig:mulhead_rep}]{%
  \includegraphics[height=5cm,width=.49\linewidth]{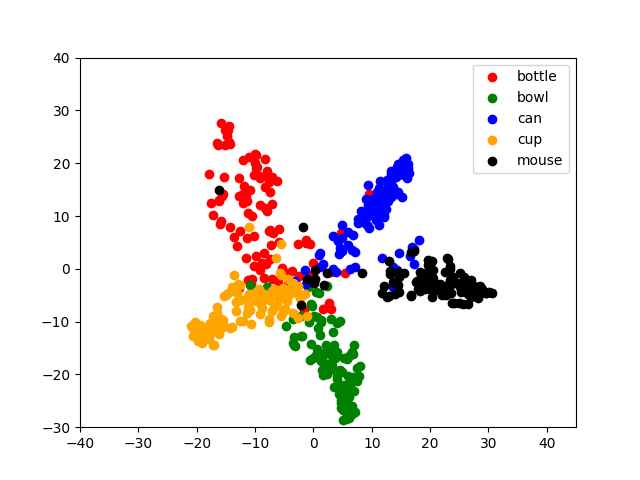}%
}
\caption{In SupCon, samples from the \textit{can} class are positioned farther from \textit{bottle}, \textit{cup}, and \textit{bowl} compared to the \textit{mouse} class. In contrast, MLCL clusters the container classes more closely in the representation space.}
\label{fig:representations4}
\end{figure}

\begin{figure}
\label{tsne-fig5}
\subfloat[SupCon\label{fig:supcon_rep}]{%
  \includegraphics[height=5cm,width=.49\linewidth]{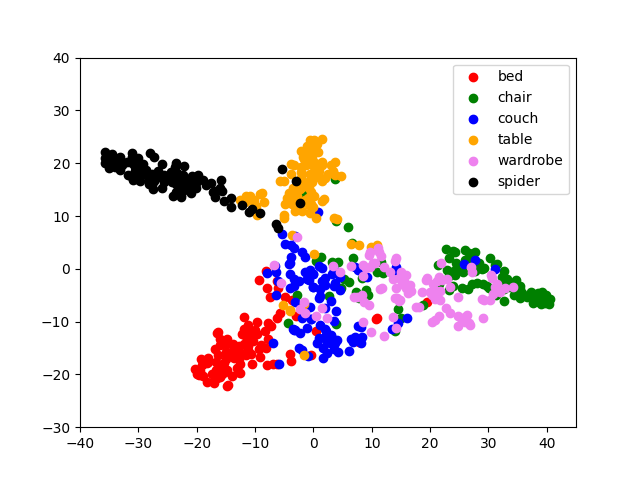}%
}
\hfill
\subfloat[MLCL\label{fig:mulhead_rep}]{%
  \includegraphics[height=5cm,width=.49\linewidth]{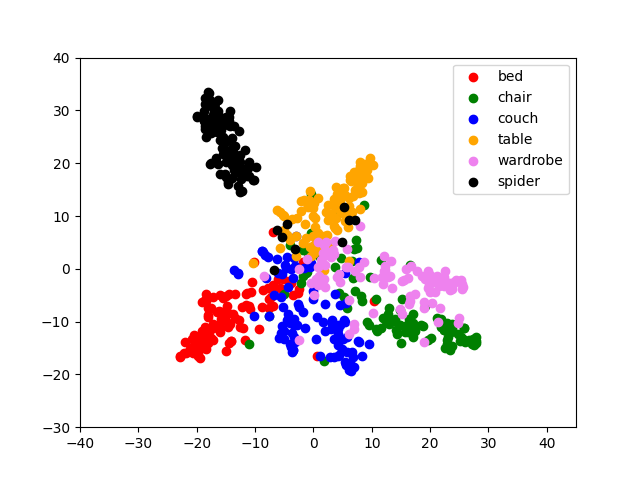}%
}
\caption{MLCL aligns the furniture classes more closely in the representation space and distinct from the class \textit{spider}.}
\label{fig:representations5}
\end{figure}

We provide additional t-SNE visualizations of the feature space for our approach compared to SupCon in Figures \ref{fig:representations2} to \ref{fig:representations5}. The second projection head plays a crucial role in preserving the proximity of representations for samples belonging to the same superclass. \newline

\vfill

\end{document}